\documentclass[sigconf]{acmart}

\usepackage{booktabs} 
\usepackage{graphicx}
\usepackage{amsfonts}
\usepackage{amsthm}
\usepackage[linesnumbered,ruled]{algorithm2e} 
\usepackage{amsfonts}
\usepackage{amsbsy}
\usepackage{listings}
\usepackage{multirow}
\usepackage{siunitx}
\usepackage{bm}
\usepackage{subfigure}
\graphicspath{{images/}{../images/}}

\usepackage{amsmath}

\newcommand{\route}{\mathcal{R}}

\newcommand{\budget}{budget}

\newcommand{\GG}{\mathcal{G}}

\newcommand{\V}{\mathcal{V}}
\newcommand{\E}{\mathcal{E}}
\newcommand{\adj}{\mathcal{A}}
\newcommand{\vect}[1]{\mathbf{#1}}

\setcopyright{acmcopyright}

\copyrightyear{2021}
\acmYear{2021}
\setcopyright{acmcopyright}\acmConference[KDD '21]{Proceedings of the 27th ACM
SIGKDD Conference on Knowledge Discovery and Data Mining}{August 14--18,
2021}{Virtual Event, Singapore}
\acmBooktitle{Proceedings of the 27th ACM SIGKDD Conference on Knowledge
Discovery and Data Mining (KDD '21), August 14--18, 2021, Virtual Event, Singapore}
\acmPrice{15.00}
\acmDOI{10.1145/3447548.3467423}
\acmISBN{978-1-4503-8332-5/21/08}

\begin{document}
\fancyhead{}
\title{Large-Scale Data-Driven Airline Market Influence Maximization}

\author{Duanshun Li}
\authornote{Duanshun Li and Jing Liu equally contributed to this work and were advised by Noseong Park when he was in George Mason University, Fairfax, VA, USA.}
\email{duanshun@ualberta.ca}
\affiliation{%
\institution{University of Alberta}
\city{Edmonton}
\state{Alberta}
\country{Canada}
}

\author{Jing Liu}
\authornotemark[1]
\email{jliu11.job@gmail.com}
\affiliation{%
\institution{Walmart Labs.}
\city{Reston}
\state{VA}
\country{USA}
}

\author{Jinsung Jeon}
\email{jjsjjs0902@yonsei.ac.kr}
\affiliation{
\institution{Yonsei University}
\city{Seoul}
\country{South Korea}
}

\author{Seoyoung Hong}
\email{seoyoungh.kr@gmail.com}
\affiliation{\institution{Yonsei University}
\city{Seoul}
\country{South Korea}
}
  
\author{Thai Le, Dongwon Lee}
\email{{thaile,dongwon}@psu.edu}
\affiliation{\institution{Penn State University}
\city{University Park}
\state{PA}
\country{USA}
}
  
\author{Noseong Park}
\email{noseong@yonsei.ac.kr}
\affiliation{\institution{Yonsei University}
\city{Seoul}
\country{South Korea}
}

\renewcommand{\shortauthors}{B. Trovato et al.}

\begin{abstract}
We present a prediction-driven optimization framework to maximize the market influence in the US domestic air passenger transportation market by adjusting flight frequencies. At the lower level, our neural networks consider a wide variety of features, such as classical air carrier performance features and transportation network features, to predict the market influence. On top of the prediction models, we define a budget-constrained flight frequency optimization problem to maximize the market influence over 2,262 routes. This problem falls into the category of the non-linear optimization problem, which cannot be solved exactly by conventional methods. To this end, we present a novel adaptive gradient ascent (AGA) method. Our prediction models show two to eleven times better accuracy in terms of the median root-mean-square error (RMSE) over baselines. In addition, our AGA optimization method runs 690 times faster with a better optimization result (in one of our largest scale experiments) than a greedy algorithm.
\end{abstract}

\ccsdesc[500]{Theory of computation~Mathematical optimization}
\ccsdesc[300]{Computing methodologies~Neural networks}

\keywords{large-scale optimization; transportation; deep learning}

\maketitle

\section{Introduction}
\begin{figure}[t!]
\centering
\includegraphics[width=1\columnwidth]{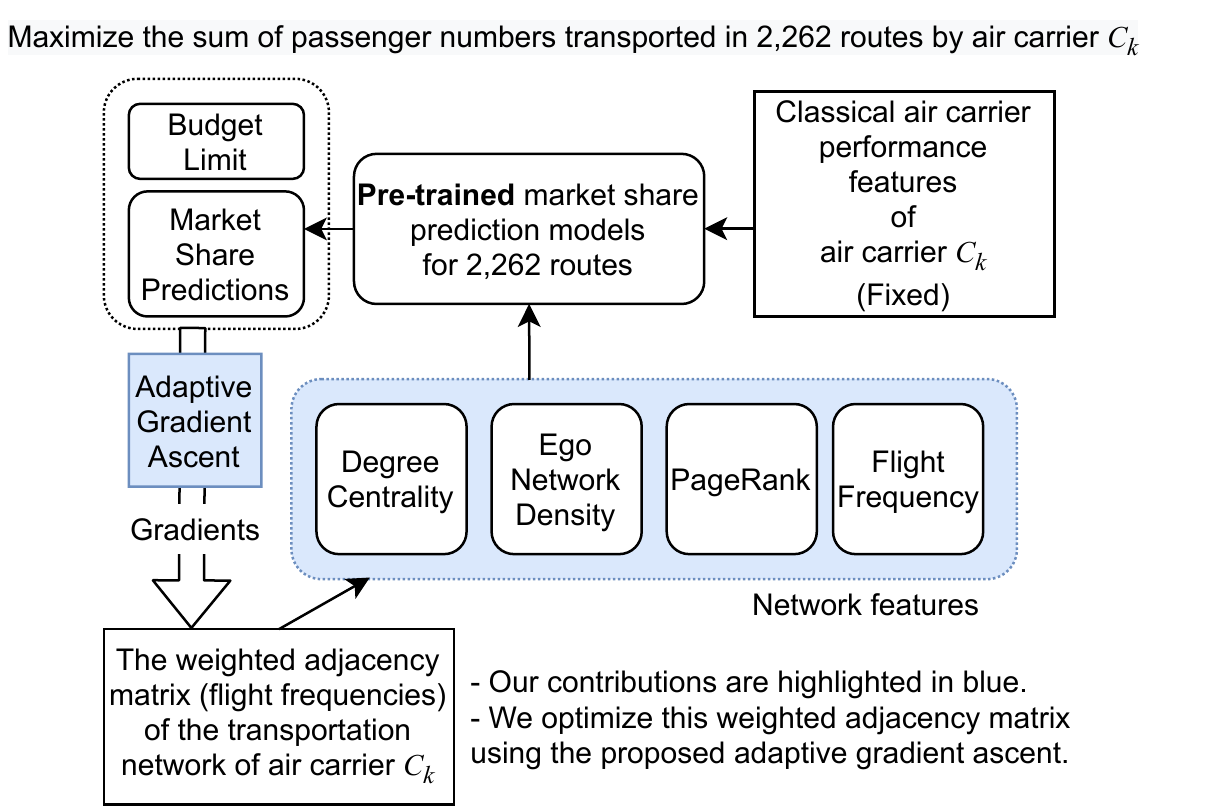}

\caption{The architecture of the proposed prediction-driven optimization framework to maximize the market influence of air carrier $C_k$. Note that the pre-trained neural network-based market share prediction models constitute the objective function. The gradients of the budget constraint and the objective function can flow from the top to the bottom to optimize the weighted adjacency matrix because all intermediate modules are differentiable.}  \label{fig:archi}
\end{figure}
Ever since the deregulation in 1978, there has been huge competition among US air carriers (airlines) for air passenger transportation. 771 million passengers were transported in 2018 alone and the largest air carrier produces a revenue of more than 43 billion dollars for the period between September 2017 and September 2018\footnote{\url{https://www.transtats.bts.gov}}. It is one of the largest domestic markets in the world and there is a huge demand to improve their services. Consequently, many computational methods have also been proposed to predict market share, ticket price, demand, etc. and allocate resources (e.g., aircraft) on those air passenger markets accordingly~\cite{An:2016:MFM:2939672.2939726,An:2017:DFA:3055535.3041217,10.1007/978-3-319-54430-4_3,8606022}.

The market influence is sometimes strategically more important than profits. Typically, there are two ways to expand business: i) a strategical merger with other strong competitors, and ii) a strategical play to maximize the market influence~\cite{ciliberto2018market}. Our paper is closely related to the latter strategy.

We propose a novel way of \emph{unifying both data mining and mathematical optimization methods} to maximize air carrier's influence on the air transportation market. In this paper, we define the influence of an air carrier as \emph{the number of passengers transported by the air carrier} which can be calculated by the total demand multiplied with the air carrier's market share.


\begin{table*}[t]
\small
\setlength{\tabcolsep}{2pt}
\begin{center}
\caption{Comparison table between two related papers~\cite{An:2016:MFM:2939672.2939726,An:2017:DFA:3055535.3041217} and this work. Since we do not assume the route independence, our problem setting is more realistic, making many existing optimization algorithms designed based on the assumption inapplicable to our work.}\label{tbl:cmp}
\begin{tabular}{|c|c|c|}
\hline

\textbf{Comparison items} & \textbf{Existing work}~\cite{An:2016:MFM:2939672.2939726,An:2017:DFA:3055535.3041217} & \textbf{Our work} \\ \hline
Market Share Prediction Model & Standard multi-logit model & Deep learning model \\ \hline
Conventional Air Carrier Performance Features & Yes & Yes \\ \hline
Transportation Network Features & No & Yes \\ \hline
Removal of Route Independence Assumption & No & Yes \\ \hline
Optimization Technique & Classical combinatorial optimization techniques & Our proposed adaptive gradient ascent \\ \hline
How to integrate prediction and optimization & Black-box query to prediction model & \begin{tabular}[c]{@{}c@{}}White-box search\end{tabular} \\ \hline
\end{tabular}
\end{center}
\end{table*}

Since the market influence of an air carrier in a route can be calculated by the total demand (passenger numbers) in the route multiplied with the market share, predicting market share is a key step in our work. Conventional features (e.g., average ticket price, flight frequency, and on-time performance) have been widely used to predict the market share~\cite{An:2016:MFM:2939672.2939726,An:2017:DFA:3055535.3041217,suzuki2000relationship,wei2005impact}. For instance, air carrier's market share on a route will increase if ticket price is decreased and flight frequency is increased. However, some researchers recently paid an attention to air carrier's transportation network connectivity that is highly likely to be connected to market share~\cite{10.1007/978-3-642-21786-9_61,doi:10.1057/ejis.2010.11}. As a response, we design a neural network-based prediction model that uses a wide variety of conventional and transportation network features, such as degree centrality, PageRank, and so forth. It is worth mentioning that we train a prediction model for each route.

On top of the market share prediction models, we build a budget-constrained optimization module to maximize the market influence by optimizing transportation network (more precisely, flight frequency values over 2,262 routes), which is an Integer Knapsack problem (cf. Fig.~\ref{fig:archi}). Our objective function consists of the market share prediction models in those routes and our constraint is a budget limit of an air carrier. The objective is not in a simple form but rather a complex one of inter-correlated neural networks because changing frequency in a route will influence market shares on other neighboring routes as well. Therefore, it is very hard to solve with existing techniques that assume routes are independent from each other (see discussions in Section~\ref{sec:opt}).

We test our optimization framework with 2,262 routes. To achieve such a high scalability, we design a method of \underline{\textbf{A}}daptive \underline{\textbf{G}}radient \underline{\textbf{A}}scent (AGA). In our experiments, the proposed optimization method solves the very large-scale optimization problem much faster than existing algorithms. However, one main challenge in our approach is how to consider the budget constraint in the proposed gradient-based optimization technique  --- each air carrier has a limited budget to operate flights. It is not straightforward to consider the budget constraint with gradient-based optimization methods. However, our proposed AGA method is able to dynamically manipulate gradients to ensure the budget limit, i.e., dynamically impose a large penalty, if any cost overrun, in such a way that one gradient ascent update theoretically guarantees a decrease in the total cost. Therefore, a series of updates can eventually address the cost overrun problem.


In our experiments, our customized prediction model shows much better accuracy in many routes than existing methods. In particular, our median root-mean-square error is more than two times better than the best baseline. Our proposed AGA method is able to maximize the market influence on all those routes 690 times faster with a better optimized influence than a greedy algorithm.


\begin{figure}[t]
\centering
\subfigure[Existing Black-box Search Methods]{\includegraphics[width=0.95\columnwidth]{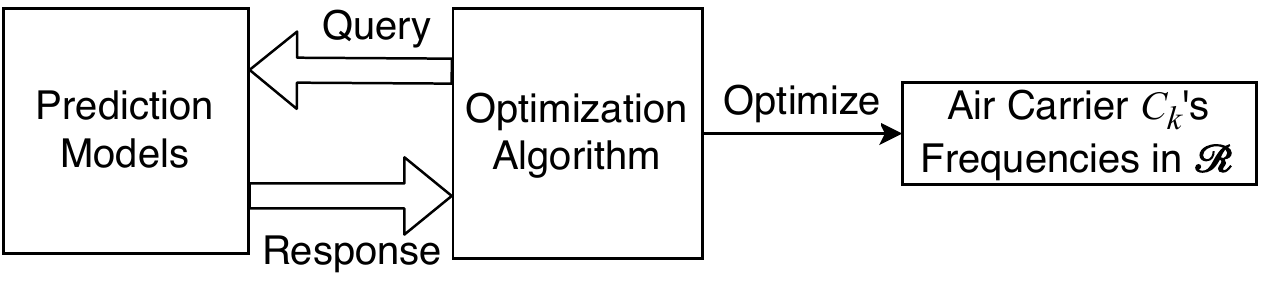}}
\subfigure[Proposed White-box Search Method]{\includegraphics[width=0.95\columnwidth]{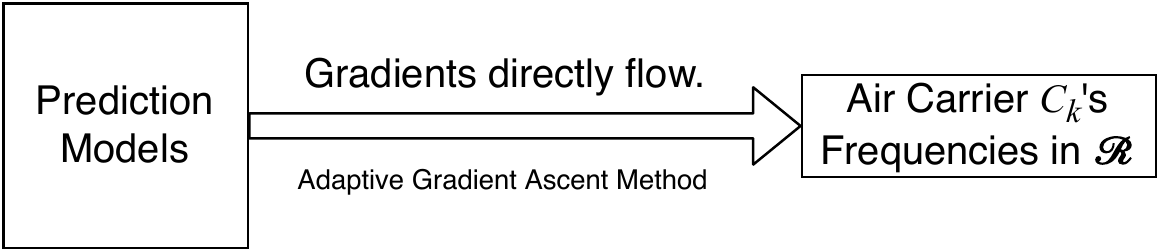}}

\caption{The comparison with existing black-box search methods in~\cite{An:2016:MFM:2939672.2939726,An:2017:DFA:3055535.3041217} and the white-box search method proposed in this work. Our AGA optimization algorithm enables the white-box search concept to be used in this work.}  \label{fig:ws}
\end{figure}

\section{Related Work}
We introduce a selected set of related works about air market predictions and optimizations.

\subsection{Market Share Prediction}
There have been proposed many prediction models such as~\cite{suzuki2000relationship,wei2005impact,An:2016:MFM:2939672.2939726,An:2017:DFA:3055535.3041217}, to name a few. However, they share many common design points. First, almost all of them use the multi-logit regression model. It is a standard model to predict air transportation market shares. We also use the same multi-logit regression (see Section~\ref{sec_pred} for its details) after some extensions. Suzuki considers air carriers' frequency, delay, and safety~\cite{suzuki2000relationship} whereas Wei et al. study about the effect of aircraft size and seat availability on market share and consider other variables such as price and frequency~\cite{wei2005impact}. There are some more similar works~\cite{An:2016:MFM:2939672.2939726,An:2017:DFA:3055535.3041217}. In our paper, we consider transportation network features in addition to those conventional air carrier performance features.


\subsection{Flight Frequency Optimization}\label{sec:opt}
One similar flight frequency optimization problem to maximize profits was solved in~\cite{An:2016:MFM:2939672.2939726,An:2017:DFA:3055535.3041217}. In their work, An et al. showed that the frequency-market share curve is very hard to approximate with existing approximation methods such as piece-wise linear approximation~\cite{646812}. After that, they designed one heuristic-based algorithm, called GroupGreedy, which runs an exact algorithm in each subset of routes (because running the exact algorithm for the entire route set is prohibitive). Each subset consists of a few routes and running the exact algorithm within a small subset provides a tolerable degree of scalability in general. However, they were able to test with \emph{at most about 30 routes} for its prohibitively long execution time even with GroupGreedy and its scalability is not satisfactory. We test with 2,262 routes in this paper --- i.e., the problem search space size is $\mathcal{O}(n^{30})$ in their work vs. $\mathcal{O}(n^{2,262})$ in this work.

In addition, we found that GroupGreedy cannot be used for our prediction model because of the network features --- An et al. did not consider network features and assumed each route is independent~\cite{An:2016:MFM:2939672.2939726,An:2017:DFA:3055535.3041217}. After adopting the assumption, they optimize for each route separately. In reality, however, changing a frequency in a route is likely to influence the market shares in other routes because routes are often inter-correlated. Thus, GroupGreedy based on the independence assumption is not applicable to our work. Our work does not assume the independence so this work is more realistic.

In the perspective of Knapsack, after excluding the independence assumption, it becomes much more complicated because the value (i.e., market share) of a product (i.e., route) becomes non-deterministic and is influenced by other products (i.e., routes). This makes the current problem more realistic than those studied in the previous work by An et al. However, this change prevents us from applying many existing Knapsack algorithms that have been invented for the simplest case where product values are fixed and independent from each other~\cite{axiotis_et_al:LIPIcs:2019:10595}.

One more significant difference is that the optimization algorithm in the related work queries its prediction models whereas both optimization and prediction are integrated on TensorFlow in this new paper. In Table~\ref{tbl:cmp}, we summarize the differences between the previous work and our work. In addition, Fig.~\ref{fig:ws} compares their fundamental difference on the algorithm design philosophy. Those existing methods are representative black-box search methods where the query-response strategy is adopted. In this new work, however, the gradients directly flow to update frequencies so its runtime is inherently faster than existing methods.

\section{Preliminaries}
We introduce our dataset and the state-of-the-art market share prediction model. Our main dataset is the air carrier origin and destination survey (DB1B) dataset released by the US Department of Transportation's Bureau of Transportation Statistics (BTS)~\cite{bts} and some safety dataset by the National Transportation Safety Board (NTSB)~\cite{ntsb}. We refer to Appendix for detailed dataset information.

\subsection{Market Share Prediction Model}\label{sec_pred}
In this subsection, we describe a popular existing market share prediction model for air transportation markets. Given a route $r$, the following multinomial logistic regression model is to predict the market share of air carrier $C_k$ in the route:
\begin{align}\label{eq:logit}{\color{black}
m_{r,k} = \frac{ e^{\sum_{j} w_{r,j} \cdot f_{r,k,j}}}{\sum_{i} e^{\sum_{j} w_{r,j} \cdot f_{r,i,j}}} = \frac{exp(\mathbf{w}_r \cdot \mathbf{f}_{r,k})}{\sum_i exp(\mathbf{w}_r \cdot \mathbf{f}_{r,i})},}
\end{align}where $m_{r,k}$ means the market share of air carrier $C_k$ in route $r$; $f_{r,k,j}$ is the $j$-th feature of air carrier $C_k$ in route $r$; and $w_{r,j}$ represents the sensitivity of market share to feature $f_{r,k,j}$ in route $r$ that can be learned from data.

A set of features for air carrier $C_k$ in route $r$ can be represented by a vector $\mathbf{f}_{r,k}$ (see Appendix~\ref{sec:features} for a complete list of $\mathbf{f}_{r,k}$ in our work). We use bold font to denote vectors.

The rationale behind the multi-logit model is that $exp(\mathbf{w}_r \cdot \mathbf{f}_{r,k})$ can be interpreted as passengers' valuation score about air carrier $C_k$ and the market share can be calculated by the normalization of those passengers' valuation scores --- this concept is not proposed by us but widely used for the air carrier market share prediction in Business, Operations Research, etc~\cite{hansen1990airline,An:2016:MFM:2939672.2939726,An:2017:DFA:3055535.3041217,suzuki2000relationship,wei2005impact}. 



\section{Proposed Prediction Method}
We design a neural network-based market share prediction model with transportation network features.

\subsection{Air Carrier Transportation Network}
There are more than 2,000 routes (e.g., from LAX to JFK) in the US and this creates one large transportation network. Transportation network $\GG=(\V,\E)$ is a directed graph among airports (i.e., vertices) in $\V$. In particular, we are interested in an air carrier-specific directed transportation network $\GG_k$ weighted by its flight frequency values. Thus, $\GG_k$ represents the connectivity of air carrier $C_k$ and its edge weight on a certain directional edge means the flight frequency of the air carrier in the route. $\GG_k$ can be represented by a weighted adjacency (or frequency) matrix $\adj_k$, where each element is a flight frequency from one airport to another.


\subsection{Network Features}\label{sec:netf}
In this section, we introduce the network features we added to improve the prediction model.

\begin{figure}[t]
\centering
\footnotesize
\includegraphics[width=0.95\columnwidth,trim={2cm 1cm 2cm 2cm},clip]{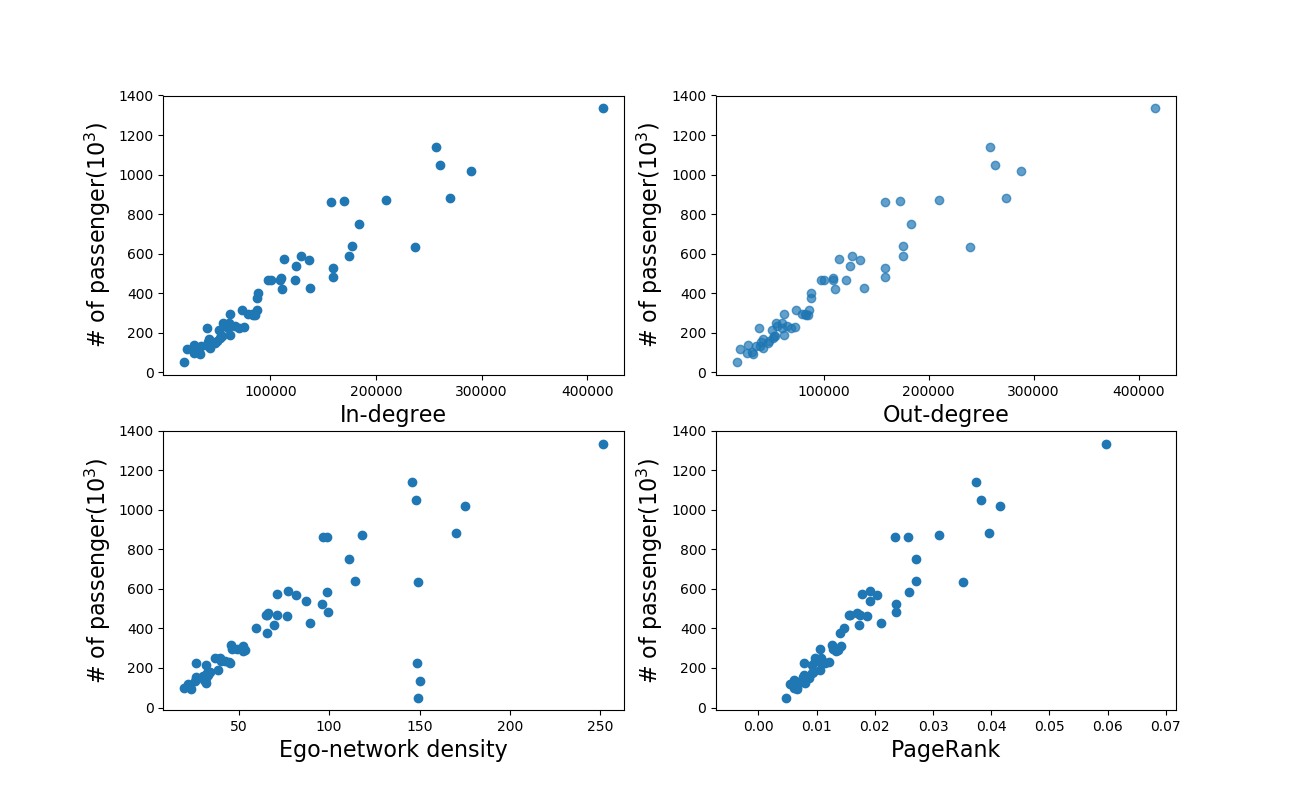}
\caption{Number of passengers vs. network features. The result summarizes all the airports.}
\label{fig:netfeat_passenger}
\end{figure}

\subsubsection{Degree Centrality}
As mentioned by earlier works, transportation network connectivity is important in air transportation markets~\cite{10.1007/978-3-642-21786-9_61,doi:10.1057/ejis.2010.11}. For instance, the higher the degree centrality of an airport in $\GG_k$, the more options the passengers to fly. Thereby, its market share will increase at the routes departing the high degree centrality airport. Therefore, we study how the degree centrality of source and destination airports influences the market share.

Given $\adj_k$, the out-degree (resp. in-degree) centrality of $i$-th airport is the sum of $i$-th row (resp. column). So this feature calculation can be very easily implemented on Tensorflow or other deep learning platforms.


\subsubsection{Ego Network Density}
Ego network is very popular for social network analysis~\cite{NIPS2012_4532}. We introduce the concept of ego network first.

\begin{definition}
Given a vertex $v$, its \emph{ego network} is an induced subgraph of $v$ and its neighbors. The vertex $v$ is called \emph{ego vertex} (i.e., ego airport in our case). Note that ego networks are also weighted with flight frequency values. The density of an ego network is defined as the sum of edge weights divided by $n(n-1)$ where $n$ is the number of vertices in the ego network.
\end{definition}

By the definition, an airport's ego network density is high when the airport and its neighboring airports are well connected all together. It is natural that passengers transit in an airport whose connections are well prepared for their final destinations. 


\subsubsection{PageRank}
PageRank was originally proposed to derive a vertex's importance score based on the random web surfer model~\cite{Page99thepagerank} --- i.e., a web surfer performs a random walk following hyperlinks. We think PageRank is suitable to analyze multi-stop passengers for the following reason.

After normalizing $\adj_k$ row-wise, it becomes the transition probability that a random passenger will move following the route. Thus, PageRank is able to capture the importance of an airport.

Fig.~\ref{fig:netfeat_passenger} depicts the relationships between the network features introduced above and the total number of passengers transported in and out airports by a certain air carrier. We used the DB1B data released by the BTS for the first quarter of 2018 to draw this figure. As shown in Fig.~\ref{fig:netfeat_passenger}, the number of passengers in each airport is highly correlated with the network features (i.e., in-degree, out-degree, ego network density, and PageRank). In conjunction with other classical air carrier performance features, these network features can improve the prediction accuracy by a non-trivial margin.

\subsection{Neural Network-based Prediction}\label{sec:nnmodel}
Whereas many existing methods rely on classical machine learning approaches, we use the following neural network to predict:
\begin{align}\begin{split}\label{eq:nn}
\mathbf{h}^{(1)}_{r,k} &= \sigma(\mathbf{f}_{r,k}\mathbf{W}^{(0)} + \mathbf{b}^{(0)}),\textrm{ for initial layer}\\
\mathbf{h}^{(i+1)}_{r,k} &= \mathbf{h}^{(i)}_{r,k} + \sigma(\mathbf{h}^{(i)}_{r,k}\mathbf{W}^{(i)} + \mathbf{b}^{(i)}),\textrm{ if }i \geq 1
\end{split}\end{align}where $\sigma$ is ReLU. $\mathbf{W}^{(0)} \in \mathcal{R}^{19 \times d}$, $\mathbf{b}^{(0)} \in \mathcal{R}^{d}$, $\mathbf{W}^{(i)} \in \mathcal{R}^{d \times d}$, $\mathbf{b}^{(i)} \in \mathcal{R}^{d}$ are parameters to learn. Note that we use residual connections after the initial layer. For the final activation, we also use the multi-logit regression. From Eq.~\eqref{eq:logit}, we replace $\mathbf{f}_{r,k}$ with $\mathbf{h}^{l}_{r,k}$, which denotes the last hidden vector of our proposed neural network, to predict $m_{r,k}$ as follows:
\begin{align}\label{eq:nn2}
m_{r,k} = \frac{exp(\mathbf{w}_r \cdot \mathbf{h}^{l}_{r,k})}{\sum_i exp(\mathbf{w}_r \cdot \mathbf{h}^{l}_{r,i})},
\end{align}where $\mathbf{w}_r$ is a trainable parameter. We use $\bm{\theta}_r$ to denote all the parameters of route $r$ in Eqs.~\eqref{eq:nn} and~\eqref{eq:nn2}.

One thing to mention is that all the network features can be properly calculated on TensorFlow from $\adj_k$ before being fed into the neural network. This is the case during the frequency optimization phase which will be described shortly. By changing a frequency in $\adj_k$, the entire network feature can be recalculated before the neural network processing as shown in Fig.~\ref{fig:archi}. Therefore, the gradients can directly flow from the prediction models to the frequency matrix through the network feature calculation part. Hereinafter, we use a function $m_{r,k}(\adj_k;\bm{\theta}_r)$ after partially omitting features (such as ticket price, aircraft size, etc.) to denote the predicted market share. Note that the omitted features and $\bm{\theta}_r$ are considered constant while optimizing frequencies in the next section. We sometimes omit all the inputs and use $m_{r,k}$ for brevity.

\section{Proposed Optimization Method}

{\color{black} Among many features, the flight frequency is an actionable feature that we are interested in to adjust --- see Appendix~\ref{sec:features} for a complete list of features we consider in this work. An actionable feature means a feature that can be freely decided only for one's own purposes. May other features, such as delay time, safety, and so on, cannot be solely decided by an air carrier. Hereinafter, we use $f_{r,k,freq}$ to denote a flight frequency value of air carrier $C_k$ in route $r$. These frequency values among airports constitute $\adj_k$.

\subsection{Problem Definition}\label{sec:def}
We solve the following optimization problem to maximize the market influence of air carrier $C_k$ (i.e., the number of passengers transported by $C_k$) on those routes in $\route$. Given its total budget $\budget_k$, we optimize the flight frequency values of the air carrier over multiple routes in $\route$ as follows:

\begin{align}\begin{split}\label{eq:obj}
\max_{f_{r}^{max} \geq f_{r,k,freq}\geq 0, r \in \route}& \sum_{r \in \route} demand_r \times m_{r,k} \\
\textrm{subject to }& \sum_{r \in \route}cost_{r,k} \times f_{r,k,freq} \leq \budget_k,
\end{split}\end{align} where $m_{r,k}$ is the predicted market share of $C_k$ in route $r$ (by our neural network model), $demand_r$ is the number of total passengers in route $r$ from the DB1B dataset, and $cost_{r,k}$ is the unit operational cost of air carrier $C_k$ in route $r$. $f_{r}^{max}$ is the maximum flight frequency in route $r$ observed in the DB1B dataset. The adoption of $f_{r}^{max}$ is our heuristic to prevent overshooting a practically meaningful frequency limit. Note that different air carriers have different unit operational costs in a route $r$ as their efficiency is different and they purchase fuel in different prices --- we extract this information from the DB1B dataset.

Eq.~\eqref{eq:obj} shows how we can effectively merge data mining and mathematical optimization. The proposed problem is basically a non-linear optimization and a special case of Integer Knapsack and resource allocation problems which are all NP-hard~\cite{Arora:2009:CCM:1540612}. The theoretical complexity of the problem is $\mathcal{O}(\prod_{r\in \route}f_{r}^{max})$, which can be simply written as $\mathcal{O}(n^{2,262})$ after assuming $n=f_{r}^{max}$ in each route for ease of discussion because $|\route|=2,262$.

\begin{theorem}
The market influence maximization is NP-hard.
\end{theorem}

\subsection{Overall Architecture}

In Fig.~\ref{fig:archi}, the overall architecture of the proposed optimization idea is shown. The overall workflow is as follows:

\begin{enumerate}
\item Train the market share prediction model in each route, which considers transportation network features.
\item Fix the prediction models and update the frequency matrix $\adj_k$ using the proposed AGA optimizer. We consider other features (such as ticket price, aircraft size, etc.) are fixed while optimizing frequencies.
\end{enumerate}

The adoption of network features makes many classical combinatorial optimization techniques inapplicable to our work because the route independent assumption does not hold any more. Even worse, our objective function consists of highly non-linear neural networks. Therefore, our problem becomes a challenging non-linear optimization problem. We shortly describe how to solve such a large-scale and difficult optimization problem.

\subsection{Gradient-based Optimization}\label{sec:sol}
We solve the problem in Eq.~\eqref{eq:obj} on a deep learning platform using our AGA method in Algorithm~\eqref{alg:adaptive-gd}. But one problem in this approach is how to consider the budget constraint. We design two workarounds based on i) Lagrangian function (LF) and ii) rectified linear unit (ReLU).\medskip

In our heuristic, we covert integer frequency variables to real variables and use the \texttt{clip\_by\_value} function of TensorFlow to restrict the frequency in $r$ into $[0,f_{r}^{max}]$ during the optimization process. As the optimized frequencies by our method will be real numbers, \emph{we round down to convert them to integers} and not to violate the budget limit at the end of the optimization process i.e.,  a continuous relaxation from integer frequencies. We now describe how to solve the continuous-relaxed problem.

\subsubsection{Lagrangian Function (LF)-based Heuristic: }
{\color{black}The method of Lagrange multiplier is a popular method to maximize concave functions (or some special non-concave functions) with constraints~\cite{10.5555/993483,10.1561/2200000016}. However, we cannot apply the method to our work because our objective function consists of highly non-linear neural networks. Therefore, we adopt only the Lagrangian function from the method and develop our own heuristic search method.} The following Lagrangian function can be defined in our case:
\begin{align}\label{eq:lag_l}\begin{split}
L = o(\adj_k) - \lambda c(\adj_k),
\end{split}\end{align} where $\lambda$ is called a Lagrange multiplier, and
\begin{align}\begin{split}
o(\adj_k) &= \sum_{r \in \route} demand_r \times m_{r,k},\\
c(\adj_k) &= \sum_{r \in \route}\big(cost_{r,k} \times f_{r,k,freq}\big) - \budget_k.
\end{split}\end{align}

{\color{black}Basically, the Lagrange multiplier $\lambda$ can be systematically decided, if the objective function $o(\adj_k)$ is in simple forms, and we can find the optimal solution of the original constrained problem. However, this is not the case in our work due to the complicated nature of neural networks and the objective function from them, and our goal is to solve the optimization problem on TensorFlow for the purpose of increasing scalability, aided by our scalable AGA optimization technique.
Thus, we propose the following regularized problem and develop a heuristic search method}:
\begin{align}\label{eq:lag2}
\max_{f_{r}^{max} \geq f_{r,k,freq}\geq 0, r\in \route}\quad \min_{\lambda}\quad L + \delta \lambda^2
\end{align}where $\delta \geq 0$ is a weight for the regularization term. {\color{black}Note that our definition is different from the original Lagrangian function.} The inner minimization part has been added by us to prevent that $\lambda$ becomes too large. One way to solve Eq.~\eqref{eq:lag2} is to alternately optimize flight frequencies (i.e., the outer maximization) and $\lambda$ (i.e., the inner minimization), which implies that Eq.~\eqref{eq:lag2} be basically a two-player max-min game. We further improve Eq.~\eqref{eq:lag2} and  derive a simpler but equivalent formulation that does not require the alternating maximization and minimization shortly in Eq.~\eqref{eq:lag_max_min_simp}.

\begin{theorem}\label{th:optimization}
Let $\adj_k$ be a matrix of flight frequencies. The optimal solution of the max-min problem in Eq.~\eqref{eq:lag2} is the same as the optimal solution of the following problem:
\begin{align}\label{eq:lag_max_min2}
    \max_{f_{r}^{max} \geq f_{r,k,freq}\geq 0, r\in \route}\quad  o(\adj_k) -\frac{c(\adj_k)^2}{4\delta}.
\end{align}
\end{theorem}

For simplicity, let $\beta = \frac{1}{2\delta}$ and we can rewrite Eq.~\eqref{eq:lag_max_min2} as follows:
\begin{align}\label{eq:lag_max_min_simp}
    \max_{f_{r}^{max} \geq f_{r,k,freq}\geq 0, r\in \route}\quad  \bar{L}_{Lagrange},
\end{align}where $\bar{L}_{Lagrange} = o(\adj_k) -\beta\frac{c(\adj_k)^2}{2}$.

Note that maximizing Eq.~\eqref{eq:lag_max_min_simp} is equivalent to solving the max-min problem in Eq.~\eqref{eq:lag2} so we implement only Eq.~\eqref{eq:lag_max_min_simp} and optimize it using the proposed AGA method that will be described in the next subsection.


\subsubsection{Rectified Linear Unit (ReLU)-based Heuristic: }
ReLU is used to rectify an input value by taking its positive part for neural networks. This property can be used to impose a penalty if the budget limit constraint is violated as follows:
\begin{align}\label{eq:lag_max_min_simp2}
\max_{f_{r}^{max} \geq f_{r,k,freq}\geq 0, r\in \route}\quad  \bar{L}_{ReLU},
\end{align}where $\bar{L}_{ReLU} = o(\adj_k) - \beta R(c(\adj_k))$ and $R(\cdot)$ is the rectified linear unit.



\subsection{$\beta$ Selection and Adaptive Gradient Ascent}
{\color{black}We propose the AGA method, which basically uses the gradients of $\bar{L}_{Lagrange}$ or $\bar{L}_{ReLU}$ w.r.t. flight frequencies to optimize them. In both methods, the coefficient $\beta$ needs to be \emph{dynamically} adjusted to ensure the budget limit rather than being fixed to a constant. For example, one gradient ascent update will increase flight frequencies even after a cost overrun if $\beta$ is not large enough. Whenever there is any cost overrun, $\beta$ should be set to such a large enough value that the total cost is decreased. 
	
\begin{figure}[t]
\centering
\footnotesize
\subfigure[$\beta=1$ is not enough to decrease cost]{\includegraphics[width=0.46\columnwidth]{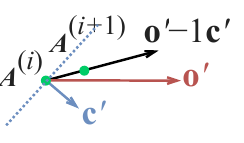}}\hfill
\subfigure[$\beta=5$ is enough to decrease cost]{\includegraphics[width=0.45\columnwidth]{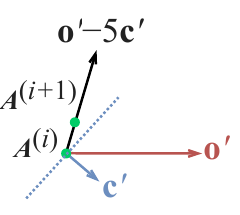}}
\caption{Suppose that there is a small cost overrun with $\adj^{(i)}$, which denotes a frequency matrix at $i$-th gradient ascent iteration. The norm of $\vect{c}'$ is smaller than that of $\vect{o}'$ and the gradient ascent update cannot remove the cost overrun if $\beta$ is small (e.g., $\beta=1$ in (a)). However, if $\beta$ is large enough (e.g., $\beta=5$ in (b)), the gradient ascent update can reduce the cost overrun. Note that $\adj^{(i+1)}$ is located behind $\adj^{(i)}$ w.r.t. the blue dotted line perpendicular to $\vect{c}'$ in (b), which means a reduced cost overrun. We dynamically adjust $\beta$ to decrease the cost, if any cost overrun, while sacrificing the objective as little as possible.}
\label{fig:ga}
\end{figure}
	
For the sake of our convenience, we will use $\vect{o}'$ and $\vect{c}'$ to denote the gradients of objective and cost overrun penalty term as follows:
\begin{align*}
\vect{o}' &= \nabla o(\adj_k),\\
\vect{c}' &= \begin{cases}\nabla\frac{c(\adj_k)^2}{2},\textrm{ if the Lagrangian function-based method}\\
\nabla R(c(\adj_k)),\textrm{ if the ReLU-based method}.
\end{cases}
\end{align*}
Fig.~\ref{fig:ga} shows an illustration of why we need to adjust $\beta$. As shown, if the directions of the two gradients $\vect{c}'$ and $\vect{o}'-\beta \vect{c}'$, where $\beta=5$, are opposite, the cost overrun will decrease after one gradient ascent update. If $\beta$ is too small, the cost overrun does not decrease in the example.
	}

	We also do not distinguish between $\bar{L}_{Lagrange}$ and $\bar{L}_{ReLU}$ in this section because the algorithm proposed in this section is commonly applicable to both the Lagrangian function and ReLU-based methods. We denote them simply as $\bar{L}$ in this section.
	
	The gradients of $\bar{L}$ w.r.t. $\adj_k$ are made of two components $\vect{o}'$ and $-\beta\vect{c}'$, where $\vect{o}'$ increases the market influence and $-\beta\vect{c}'$ reduces the cost overrun. Typically, the market influence increases as the frequencies $\adj_k$ increase. So $\beta$ needs to be properly selected such that the frequencies are updated (by the proposed AGA method) to reduce the cost once the total cost exceeds the budget during the gradient-based update process.
	This requires that the overall gradients $\vect{o}'-\beta \vect{c}'$ suppresses an increase in $c(\adj_k)$.
	More precisely, it requires that the directional derivative of $c(\adj_k)$ along the vector $\vect{o}'-\beta \vect{c}'$ (or the dot product of $\vect{o}'-\beta\vect{c}'$ and $\vect{c}'$) is negative --- if two vectors have different directions, their dot product is negative.
	
	
	    	\begin{algorithm}[t!]
		\SetAlgoLined
		\caption{Adaptive gradient ascent (AGA)}\label{alg:adaptive-gd}
		\KwIn{$\gamma$} 
		\KwOut{$\adj_k$}
		Initialize $\adj_k$\tcc*[r]{Initialize freqs}
        $\beta \gets 0$\tcc*[r]{Initialize $\beta$}
		\While {until convergence}{ 
			$\adj_k \gets \adj_k+\gamma \nabla \bar{L}$\tcc*[r]{Gradient ascent}\label{alg:opt}
			\eIf{$c(\adj_k)>0$}{
				$\beta \gets$Eq.~\eqref{eq:betafinal} 
			}{
			$\beta \gets 0$\;
			}
		}
	\end{algorithm}
	
	
Therefore, we want $\vect{c}' \cdot (\vect{o}'-\beta \vect{c}') < 0$.	From it, we can rewrite the inequality w.r.t. $\beta$ and we have
	\begin{align}\label{th:beta_selection}
	    \beta > \frac{\vect{c}' \cdot \vect{o}'}{\vect{c}' \cdot \vect{c}'}.
	\end{align}
	
    
    Note that Eq.~\eqref{th:beta_selection} does not include the equality condition but requires that $\beta$ is strictly larger than its right-hand side. To this end, we introduce a positive value $\epsilon>0$ as follows: 
    \begin{align}
    \beta = \frac{\vect{o}'\cdot\vect{c}'}{\vect{c}'\cdot\vect{c}'}+\epsilon,
    \end{align}where $\epsilon$ is a positive hyper-parameter in our method.
    
    On the other hand, we need to ensure that $\beta$ is getting closer to zero when the algorithm is approaching an optimal solution of $\adj_k$. To do this, we further modify it as follows:
    \begin{align}\label{eq:betafinal}
    \beta = \frac{\vect{o}'\cdot\vect{c}'}{\vect{c}'\cdot\vect{c}'}+c(\adj_k)\epsilon.
    \end{align}
    
    Note that $c(\adj_k)\epsilon$ becomes a very trivial value if $c(\adj_k)$ is very small. This specific setting prevents the situation that an ill-chosen large $\epsilon$ decreases flight frequencies too much given a very small cost overrun $c(\adj_k) \approx 0$.
    
    The proposed AGA method is presented in Algorithm \ref{alg:adaptive-gd}. The optimization of frequencies occurs at line~\ref{alg:opt} and other lines are for dynamically adjusting $\beta$. We take a solution around 500 epochs when the cost overrun is not positive. 500 epochs are enough to reach a solution point in our experiments.
    
    {\color{black}
\begin{theorem}
Algorithm~\ref{alg:adaptive-gd} is able to find a feasible solution of the original problem in Eq.~\eqref{eq:obj}.
\end{theorem}

}

\section{Experiments}
In this section, we introduce experimental environments and results for both the prediction and the optimization. {\color{black}We collected our data for 10 years from the website~\cite{bts}. We predict the market share and optimize the flight frequency in the last month of the dataset after training with all other month data.}

In our dataset, there are 2,262 routes and more than 10 air carriers. We predict and optimize for the top-4 air carriers among them considering their influences on the US domestic air markets. We ignore other regional/commuter level air carriers.

Our detailed software and hardware environments are as follows: Ubuntu 18.04.1 LTS, Python ver. 3.6.6, Numpy ver. 1.14.5, Scipy ver. 1.1.0, Pandas ver. 0.23.4, Matplotlib ver.3.0.0, Tensorflow-gpu ver. 1.11.0, CUDA ver. 10.0, NVIDIA Driver ver. 417.22. Three machines with i9 CPU and GTX1080Ti are used.

\subsection{Market Share Prediction}

\subsubsection{Baseline Methods}
We compare our proposed model with two baseline prediction models. 
Model1~\cite{suzuki2000relationship} considers air carrier's frequency, delay, and safety. Model2~\cite{wei2005impact} studies the effect of aircraft size and seat availability on market share and considers all other variables such as price and frequency. Model1 and Model2 are conventional methods based on multi-logit regression and they are trained using numerical solvers. Model3 is a neural network-based model created by us and uses the network features as well.

{\color{black}To train the market share prediction models, we use the learning rate of 1e-4 which decays with a ratio of 0.96 every 100 epochs. The number of layer in our neural network is $l=\{3,4,5\}$ and the dimensionality of the hidden vector is $d=\{16, 32\}$. We train 1,000 epochs for each model and use the Xavier initializer~\cite{Glorot10understandingthe} for initializing weights and the Adam optimizer for updating weights. We used the cross validation method to choose the best one, which means given a training set with $N$ months, we choose a random month and validate with the selected month after training with all other $N-1$ months. We repeat this $N$ times.

In addition, we test other standard regression algorithms as well. In particular, we are interested in testing some robust regression algorithms such as TheilSen, AdaBoost Regression, and RandomForest Regression. We also use the same cross validation method.}

\begin{figure}\centering
\footnotesize
\includegraphics[width=1\columnwidth]{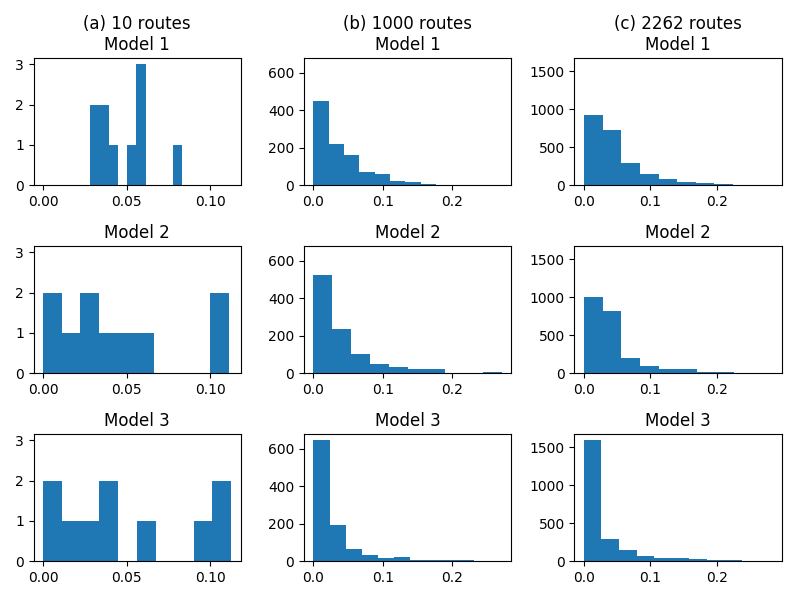}
\caption{Histogram of RMSE scores --- lower values are preferred. X-axis is the RMSE score and Y-axis is the number of routes.}
\label{fig:pred_baseline}
\end{figure}

\begin{table}\centering
\footnotesize
\caption{Median/Average RMSE and $R^2$. The up-arrow (resp. down-arrow) means higher (resp. lower) is better. The best results are indicated in bold font.}
\begin{tabular}{@{}cc|ccc@{}}\hline
	&&Median RMSE $\downarrow$	&$R^2\uparrow$	&Mean RMSE $\downarrow$\\\hline
	\multirow{6}{*}{10 routes}
	&TheilSen	&0.048	&0.944	&0.052\\
&AdaBoost   	&0.029	&0.970	&0.027\\
&RandomForest  	&0.029	&\textbf{0.979}	&0.025\\
&Model1~\cite{suzuki2000relationship}	&0.026	&0.965	&\textbf{0.024}\\
&Model2~\cite{wei2005impact}   	&0.035	&0.953	&0.030\\ 
&{\bf Model3 (Ours)}   	&\textbf{0.023}	&0.899	&0.026\\
&Model3 (No Net.)   	&0.026	&0.884	&0.027\\\hline
\multirow{6}{*}{1,000 routes}
&TheilSen	&0.080	&0.855	&0.087\\
&AdaBoost   	&0.021	&0.964	&0.033\\
&RandomForest  	&0.024	&0.968	&0.033\\
&Model1~\cite{suzuki2000relationship}	&0.021	&0.957	&0.033\\
&Model2~\cite{wei2005impact}   	&0.020	&0.983	&0.035\\
&{\bf Model3 (Ours)}   	&\textbf{0.010}	&\textbf{0.988}	&\textbf{0.025}\\
&Model3 (No Net.)   	&0.019	&0.978	&0.030\\\hline
\multirow{6}{*}{2,262 routes}
&TheilSen	&0.0813	&0.707	&0.088\\
&AdaBoost   	&0.017	&0.933	&\textbf{0.031}\\
&RandomForest  	&0.014	&0.942	&\textbf{0.031}\\
&Model1~\cite{suzuki2000relationship}	&0.033	&0.944	&0.041\\
&Model2~\cite{wei2005impact}   	&0.030	&0.976	&0.033\\
&{\bf Model3 (Ours)}   	&\textbf{0.007}	&\textbf{0.983}	&0.038\\
&Model3 (No Net.)   	&0.013	&0.969	&0.040\\

\hline
\end{tabular}
\label{table:avg_rmse_predbaseline}
\end{table}

\subsubsection{Experimental Results}
Fig.~\ref{fig:pred_baseline} shows the histogram of RMSE scores for Model1, 2, and 3. We experimented three scenarios (i.e., top-10 routes, top-1,000 routes, and top-2,262 routes in terms of the number of passengers). Our Model3 shows a higher density in low-RMSE regions than other models.

The median/average root-mean-square error (RMSE) and $R^2$ scores are summarized in Table~\ref{table:avg_rmse_predbaseline}. Our Model3 has much better median RMSE and $R^2$ scores than other models (especially for the largest scale prediction with 2,262 routes). Sometimes our mean RMSE is worse than other baselines. However, we think this is not significant because our low median RMSE says that it is better than others in the majority of routes. In particular, we show the median RMSE of 0.007 for the 2,262-route predictions vs. 0.030 by Model2. RandomForest also shows reasonable accuracy in many cases.

For the top-10 routes, most models have good performance. 
This is because it is not easy for our model to have reliable network features only with the 10 routes. However, our main goal is to predict accurately in a larger scale prediction.

{\color{black}We also compare the accuracy of our proposed model without the network features, denoted with ``No Net.'' in the table. When we do not use any network features, the accuracy of market share predictions slightly decreases. Considering the scale of the market size, however, a few percentage errors can result in a big loss in the optimization phase. Therefore, our proposed prediction model is the most suitable to be used to define the objective function of our proposed optimization problem.}

\subsection{Market Influence Maximization}

\subsubsection{Baseline Methods}\label{sec:base}
Dynamic programming, branch and bound, and GroupGreedy were used to solve a similar problem in~\cite{An:2016:MFM:2939672.2939726,An:2017:DFA:3055535.3041217}. However, all these algorithms assume that routes are independent, which is not the case in our work because we use the network features. Therefore, their methods are not applicable to our work (see Section~\ref{sec:opt}).

Therefore, we describe two baseline methods: greedy and an exhaustive algorithm. Greedy methods are effective in many optimization problems. In particular, greedy provides an approximation ratio of around 63\% for submodular minimization. Unfortunately, our optimization is not a submodular case. Due to its simplicity, however, we compare with the following greedy method, which iteratively chooses a route with the maximum marginal increment of market influence and increases its flight frequency by $\alpha$. In general, the step size $\alpha$ is 1. For faster convergence, however, we test various $\alpha =\{1, 10\}$. The complexity of the greedy algorithm is $\mathcal{O}( \frac{\budget_k\cdot N_k}{\alpha \cdot avg\_cost_{k}})$, where $N_k$ is the number of routes and $avg\_cost_{k}$ is the average cost for air carrier $k$ over the routes. However, this greedy is still a black-box method, whose efficiency is worse than our white-box method.

We can also use a brute-force algorithm when the number of routes is small. Given three routes $\{r_1, r_2, r_3\}$, for instance, the possible number of solutions is $f_{r_1}^{max} \times f_{r_2}^{max} \times f_{r_3}^{max}$. It is already a very large search space because each $f_{r_i}^{max}$ is several hundreds for a popular route in a month. However, we do not need to test solutions one by one. We create a large tensor of $|\route| \times |\route| \times q$ dimensions, where $q$ is the number of queries, and query $q$ solutions at the same time. In general, GPUs can solve the large query quickly. Even with GPUs, however, we cannot query more than a few routes because the search space volume exponentially grows. We also use the step size $\alpha=\{5,10\}$. $\alpha=1$ is not feasible in the brute-force search even with state-of-the-art GPUs. Thus, the complexity becomes $\mathcal{O}(\frac{f_{r_1}^{max}}{\alpha} \times \frac{f_{r_2}^{max}}{\alpha} \times \frac{f_{r_3}^{max}}{\alpha})$.

\begin{table}[t]
\centering
\footnotesize
\caption{Optimization results for the top-3 routes. Multiplying by 10 will lead to the real scale of passenger numbers because the DB1B database includes 10\% random samples of air tickets. LF and ReLU mean our Lagrangian function and ReLU-based methods, respectively.}
\begin{tabular}{@{}cc|cccc@{}}\hline            
&&Carrier&Carrier&Carrier&Carrier\\
&&1&2&3&4\\ \hline
\multirow{9}{*}{\rotatebox[origin=c]{90}{\# of Passengers}} 
&\multicolumn{1}{|c|}{Ground Truth} &4,960 &307 &1,792 &3,124\\
&\multicolumn{1}{|c|}{LF, Real\_Init (Ours)} 		  &4,964	&308	&1,842	&3,126  \\
&\multicolumn{1}{|c|}{ReLU, Real\_Init (Ours)}			  &4,961	&\textbf{310}	&1,854	&\textbf{3,144}  \\
&\multicolumn{1}{|c|}{LF, Zero\_Init (Ours)}		  &4,970	&308	&\textbf{1,891}	&3,139  \\
&\multicolumn{1}{|c|}{ReLU, Zero\_Init (Ours)} 			  &4,961	&\textbf{310}	&\textbf{1,891}	&\textbf{3,144}  \\
&\multicolumn{1}{|c|}{Greedy, Zero\_Init, $\alpha=1$}	  &4,967	&\textbf{310}	&\textbf{1,891}	&\textbf{3,144}  \\
&\multicolumn{1}{|c|}{Greedy, Zero\_Init, $\alpha=10$}	  &\textbf{4,972}	&\textbf{310}	&\textbf{1,891}	&\textbf{3,144}  \\ 
&\multicolumn{1}{|c|}{Brute-force, Zero\_Init, $\alpha=5$} & \textbf{4,972} & N/A & N/A & N/A\\
&\multicolumn{1}{|c|}{Brute-force, Zero\_Init, $\alpha=10$} & \textbf{4,972} &\textbf{310}&\textbf{1,891}&\textbf{3,144}\\\hline
\end{tabular}
\label{table:r3}
\vspace{1.5em}

\centering
\footnotesize
\caption{Optimized number of passengers for the top-10 routes.}
\begin{tabular}{@{}cc|cccc@{}}\hline            
&&Carrier&Carrier&Carrier&Carrier\\
&&1&2&3&4\\ \hline
\multirow{7}{*}{\rotatebox[origin=c]{90}{\# of Passengers}} 
&\multicolumn{1}{|c|}{Ground Truth} &16,924 &4,022 &20,064 &29,419\\
&\multicolumn{1}{|c|}{LF, Real\_Init (Ours)} &18,612	&4,054	&20,552	&30,220	\\
&\multicolumn{1}{|c|}{ReLU, Real\_Init (Ours)}	&18,618	&\textbf{5,024}	&\textbf{20,703} &\textbf{30,269}\\
&\multicolumn{1}{|c|}{LF, Zero\_Init (Ours)}&18,583 &4,259	&20,549	&30,074	\\
&\multicolumn{1}{|c|}{ReLU, Zero\_Init (Ours)} &\textbf{18,643}	&\textbf{5,024}	&20,323	&\textbf{30,269}\\
&\multicolumn{1}{|c|}{Greedy, Zero\_Init, $\alpha=1$}&17,016		&\textbf{5,024}	&20,515	&29,519			\\
&\multicolumn{1}{|c|}{Greedy, Zero\_Init, $\alpha=10$}	&18,078	&\textbf{5,024}	&20,515	&\textbf{30,269}\\ \hline
\end{tabular}
\label{table:r10}
\vspace{1.5em}

\centering
\footnotesize
\caption{Running time (in sec.) for the top-10 routes.}
\begin{tabular}{@{}cc|cccc@{}}\hline         
&&Carrier&Carrier&Carrier&Carrier\\
&&1&2&3&4\\ \hline
\multicolumn{2}{c|}{LF, Real\_Init (Ours)} &40.77	&42.52	&41.86	&41.70	\\
\multicolumn{2}{c|}{ReLU, Real\_Init (Ours)}&\textbf{40.75}	&41.48	&\textbf{40.67}	&44.30	\\
\multicolumn{2}{c|}{LF, Zero\_Init (Ours)}&43.10	&42.45	&40.94	&40.37\\
\multicolumn{2}{c|}{ReLU, Zero\_Init (Ours)}&40.98	&\textbf{39.90}	&40.49	&\textbf{40.31} \\
\multicolumn{2}{c|}{Greedy, Zero\_Init, $\alpha=1$}&910.12	&191.12	&1,074.95	&1,001.04	\\
\multicolumn{2}{c|}{Greedy, Zero\_Init, $\alpha=10$}&89.47	&20.14	&107.82	&101.01\\ \hline
\end{tabular}
\label{table:r10_time}
\end{table}

\begin{table*}
\centering
\footnotesize
\caption{Optimized number of passengers for the top-1,000 and 2,262 routes. Greedy with $\alpha=1$ is not feasible in this scale of experiments.}
\begin{tabular}{@{}c|cccccccc@{}}\hline     
&\multicolumn{2}{c}{Carrier 1}&\multicolumn{2}{c}{Carrier 2}&\multicolumn{2}{c}{Carrier 3}&\multicolumn{2}{c}{Carrier 4}\\
\cmidrule{2-9}
&1000 routes &2262 routes&1000 routes &2262 routes&1000 routes &2262 routes&1000 routes &2262 routes\\ \hline
\multicolumn{1}{c|}{LF, Real\_Init (Ours)}    &429,581	    &487,475	&\textbf{225,623}	&307,815	&388,864	&447,633	&546,742	&723,522\\
\multicolumn{1}{c|}{ReLU, Real\_Init (Ours)}	    &431,261	    &489,684	&\textbf{225,623}	&\textbf{307,881}	&\textbf{390,239}	&\textbf{448,421}	&547,623	&725,526\\
\multicolumn{1}{c|}{LF, Random\_Init (Ours)}  &426,683	    &\textbf{498,511}	&225,057	&306,268	&373,756	&435,277	&548,310	&\textbf{726,142}\\
\multicolumn{1}{c|}{ReLU, Random\_Init (Ours)} &\textbf{434,154}	& 492,784	&224,603	&306,963	&380,318	&447,656	&\textbf{549,092}	&721,100\\
\multicolumn{1}{c|}{Greedy, Zero\_Init, $\alpha=10$}&428,196&N/A&225,322&N/A&385,607&N/A&516,348&N/A\\ \hline
\end{tabular}
\label{table:opt_r1000_2262}
\end{table*}

\begin{table*}
\centering
\footnotesize
\caption{Running time (in sec.) for the top-1,000 and 2,262 routes scenarios.}
\begin{tabular}{@{}c|cccccccc@{}}\hline     
&\multicolumn{2}{c}{Carrier 1}&\multicolumn{2}{c}{Carrier 2}&\multicolumn{2}{c}{Carrier 3}&\multicolumn{2}{c}{Carrier 4}\\
\cmidrule{2-9}
&1000 routes &2262 routes&1000 routes &2262 routes&1000 routes &2262 routes&1000 routes &2262 routes\\ \hline
\multicolumn{1}{c|}{LF, Real\_Init (Ours)} &440.21	&875.15	&450.30	&964.00	&\textbf{451.10}&	{951.56}	&{439.56}	&{940.17}\\
\multicolumn{1}{c|}{ReLU, Real\_Init (Ours)}	&439.18	&{908.66}	&{452.39}	&947.35	&453.50&	\textbf{937.25} &438.75	&950.60\\
\multicolumn{1}{c|}{LF, Random\_Init (Ours)}&\textbf{438.06} &\textbf{878.73}	&451.15	&947.35	&453.39	&949.80	&440.17	&948.20\\
\multicolumn{1}{c|}{ReLU, Random\_Init (Ours)} &442.12	&891.05	&\textbf{448.85}	&\textbf{936.29}	&452.47	&967.23	&\textbf{438.49}	&\textbf{928.13}\\
\multicolumn{1}{c|}{Greedy, Zero\_Init, $\alpha=10$}&84,643.31&N/A&13,414.46&N/A&35,116.56&N/A&302,272.43&N/A\\ \hline
\end{tabular}
\label{table:time_r1000_2262}
\end{table*}

\subsubsection{Hyperparameter Setup}
For all methods, we let the flight frequency $f_{r,k,freq}$ of air carrier $C_k$ in route $r$ on or below the maximum frequency $f_{r}^{max}$ observed in the DB1B database. This  is very important to ensure feasible frequency values because too high frequency values may not be accepted in practice due to limited capacity of airports. This restriction can be implemented using the $clip\_by\_value(\cdot)$ function of Tensorflow.

In addition, we need to properly initialize frequency values in Algorithm~\ref{alg:adaptive-gd}. We test two ways to initialize frequencies: i) Real\_Init initializes the flight frequency values with the ground-truth values observed in the dataset, and ii) Zero\_Init initializes all the frequencies to zeros. In all methods, we set the total budget to the ground-truth budget.

We tested $\epsilon =\{1, 100, 1000\}$ but there is no significant difference on the achieved final optimization values. For the following experiments, we choose $\epsilon=1000$ to speed up the optimization process. We use 10 for the learning rate $\gamma$ and run 500 epochs.

One more thing is that the DB1B database includes 10\% random samples of air tickets\footnote{See the overview section in \url{https://www.transtats.bts.gov/DatabaseInfo.asp?DB_ID=125}.} so our reported passenger numbers multiplied by 10 will be the real scale. In this paper, we list values in the original scale of the DB1D database for better reproducibility.

\subsubsection{Experimental Results}
We first compare all the aforementioned methods in a small sized problem with only 3 routes. Especially, the brute-force search is possible only for this small problem.

For the top-3 route optimization, we choose the top-3 biggest routes and the top-4 air carriers in terms of the number of passengers transported and optimize the flight frequencies in the 3 routes for each air carrier for the last month of our dataset. In Table~\ref{table:r3}, detailed optimized market influence values are listed for various methods. Surprisingly, all methods mark similar values. We think all methods are good at solving this small size problem. However, the brute-force method is not feasible for some cases where the maximum frequency limits $f_{r}^{max}$ in the routes are large --- we mark with `N/A' for those whose runtime is prohibitively large.


Experimental results of the top-10 route optimization are summarized in Table~\ref{table:r10}. Our method based on the ReLU activation produces the best results for all the top-4 air carriers. Our Lagrangian function (LF)-based optimization also produces many reasonable results better than Greedy. Greedy shows the worst performance in this experiment. In Table~\ref{table:r10_time}, their runtimes are also reported. Our method is 2-22 times faster than the Greedy except Carrier 2 with $\alpha=10$.

For the top-1,000 and 2,262 routes, experimental results are listed in Tables~\ref{table:opt_r1000_2262} and~\ref{table:time_r1000_2262}. Our methods produce the best optimized value in the least amount of time. In particular, our method is about 690 times faster than the Greedy with $\alpha=10$ at Carrier 4. Greedy is not feasible for 2,262 routes.

Our method shows a \emph{(sub-)linear} increment of runtime w.r.t. the number of routes. It takes about 40 seconds for the top-10 routes and 400 seconds for the top-1,000 routes. When the problem size becomes two orders of magnitude larger from 10 to 1,000, the runtime increases only by one order of magnitude. Considering that we solve a NP-hard problem, the sub-linear runtime increment is an outstanding achievement. Moreover, our method consistently shows the best optimized values in almost all cases.

{\color{black}Greedy is slower than our method due to its high complexity $\mathcal{O}( \frac{\budget_k\cdot N_k}{\alpha \cdot avg\_cost_{k}})$ as described in Sec.~\ref{sec:base}. When the budget limit $\budget_k$ and the number of routes $N_k$ are large, it should query the prediction models many times, which significantly delays its solution search time. Therefore, Greedy is a classical black-box search method whose efficiency is much worse than our proposed method. One can consider our method as a white-box search method because the gradients flow directly to update flight frequencies.}

\section{Conclusion}
We presented a prediction-driven optimization framework for maximizing air carriers' market influence, which includes neural network-based market share prediction models by adding transportation network features and innovates large-scale optimization techniques through the proposed AGA method. Our approach suggests a way to unify data mining and mathematical optimization.
Our network feature-based prediction shows better accuracy than existing methods. Our AGA method can optimize for all the US domestic routes in our dataset at the same time whereas state-of-the-art methods are applicable to at most tens of routes.


\begin{acks}
Noseong Park is the corresponding author. This work of Jinsung Jeon, Seoyoung Hong, and Noseong Park was supported by the Institute of Information \& Communications Technology Planning \& Evaluation (IITP) grant funded by the Korea government (MSIT) (No. 2020-0-01361, Artificial Intelligence Graduate School Program (Yonsei University)). The work of Thai Le and Dongwon Lee was in part supported by NSF awards \#1909702, \#1940076, \#1934782, and \#2114824.
\end{acks}

\bibliographystyle{ACM-Reference-Format}
\bibliography{sample-bibliography}

\clearpage
\appendix
\section{Proofs}
\begin{theorem}
The proposed market influence maximization is NP-hard.
\end{theorem}
\begin{proof}
We prove the theorem by showing that an arbitrary Integer Knapsack problem instance can be reduced to a special case of our market influence maximization problem.

In an Integer Knapsack problem, there are $n$ product types and each product type $p$ has a value $v_p$ and a cost $c_p$. In particular, there exist an enough number of product instances for a product type so we can choose multiple instances for a certain product type. Given a budget $B$, we can choose as many instances as we want such that the sum of the product values are maximized.

This problem instance can be reduced to a market influence maximization by letting a product type $p$ be a route $r$, $c_p$ be $cost_{r,k}$, and $v_p$ be a deterministic increment of market influence by increasing the frequency by one.

Therefore, the proposed market influence maximization problem is NP-hard.
\end{proof}

\begin{theorem}
Let $\adj_k$ be a matrix of flight frequencies. The proposed max-min method in Eq.~\eqref{eq:lag2} is equivalent to $$\underset{f_{r}^{max} \geq f_{r,k,freq}\geq 0, r\in \route}{\max}\ o(\adj_k) -\frac{c(\adj_k)^2}{4\delta}.$$
\end{theorem}
\begin{proof}
First we rewrite Eq.~\eqref{eq:lag2} as follows:
\begin{align}\label{eq:lag_max_min}
    \max_{f_{r}^{max} \geq f_{r,k,freq}\geq 0, r\in \route}\quad \min_{\lambda}\quad o(\adj_k) - \lambda c(\adj_k) + \delta \lambda^2.
\end{align}
Let us fix $\adj_k$ then Eq.~\eqref{eq:lag_max_min} becomes a quadratic function (parabola) w.r.t. $\lambda$. It is already known that the optimal solution to minimize the quadratic function given a fixed $\adj_k$ is achieved when its derivative w.r.t. $\lambda$ is zero, i.e., $\nabla_{\lambda}{o(\adj_k) - \lambda c(\adj_k) + \delta \lambda^2} = - c(\adj_k) + 2\delta \lambda = 0$.  Therefore, the optimal form of $\lambda$ can be derived as $\hat{\lambda} = \frac{c(\adj_k)}{2\delta}$.

Let us substitute $\lambda$ for its optimal form $\hat{\lambda}$ in Eq.~\eqref{eq:lag_max_min} and the inner minimization will disappear as follows:
\begin{align}
    \max_{f_{r}^{max} \geq f_{r,k,freq}\geq 0, r\in \route}\quad  o(\adj_k) -\frac{c(\adj_k)^2}{4\delta}
\end{align}
\end{proof}





\begin{theorem}
Algorithm~\ref{alg:adaptive-gd} is able to find a feasible solution of the original problem in Eq.~\eqref{eq:obj}.
\end{theorem}
\begin{proof}
In Eq.~\eqref{th:beta_selection}, we choose a $\beta$ configuration that meets $\vect{c}' \cdot (\vect{o}'-\beta \vect{c}')<0$. The frequency matrix $\adj_k$ is updated by the gradient ascent rule, denoted $\adj_k = \adj_k + \gamma(\vect{o}'-\beta \vect{c}')$. However, the directions of $\vect{o}'-\beta \vect{c}'$ and $\vect{c}'$ are opposite to each other (because their dot-product is negative), which means the gradient ascent update will decrease the cost overrun term $c(\adj_k)$ as illustrated in Fig.~\ref{fig:ga}.

Therefore, after applying the proposed gradient ascent multiple times any cost overrun can be removed. Our algorithm stops at the first solution whose cost overrun is not positive after at least 500 epochs. Therefore, Algorithm~\ref{alg:adaptive-gd} is able to find a feasible solution that meets the budget constraint and its termination is guaranteed. 
\end{proof}

\section{Datasets}
Our main dataset is the air carrier origin and destination survey (DB1B) dataset released by the US Department of Transportation's Bureau of Transportation Statistics (BTS)~\cite{bts}. They release 10\% of tickets sold in the US every quarter of year for research purposes, in conjunction with much detailed air carrier information. Itemized operational expenses of air carrier are very well summarized in the dataset and for instance, we can know that how much each air carrier had paid for fuel and attendants and what kinds of air crafts were used by a certain air carrier in a certain route. Air carrier's performance is also one important type of information in the dataset. We also use some safety dataset by the National Transportation Safety Board (NTSB)~\cite{ntsb}. We list the links to the web pages where we downloaded our dataset.

\begin{enumerate}
    \item{\url{https://www.transtats.bts.gov/Tables.asp?DB_ID=125}}
    \begin{enumerate}
        \item DB1B is one of the main tables in the database and contains randomly sampled itineraries.
    \end{enumerate}
    \item \url{https://www.transtats.bts.gov/Tables.asp?DB_ID=110}
    \begin{enumerate}
        \item T-100 Domestic Market contains detailed information about markets (i.e., routes or segments).
    \end{enumerate}
    \item \url{https://www.transtats.bts.gov/Tables.asp?DB_ID=120}
    \begin{enumerate}
        \item Airline On-Time Performance Data contains detailed delay and cancel information about certain flights.
    \end{enumerate}
    \item \url{https://www.transtats.bts.gov/Tables.asp?DB_ID=135}
    \begin{enumerate}
        \item Air Carrier Financial Reports data contains the operational expense for most U.S. air carriers.
    \end{enumerate}
    \item{\url{https://www.ntsb.gov/_layouts/ntsb.aviation/index.aspx}}
    \begin{enumerate}
        \item The NTSB aviation accident database contains all civil aviation accident records ever since 1962.
    \end{enumerate}
\end{enumerate}

{\color{black}
\subsection{Data Crawling}
The Bureau of Transportation Statistics (BTS) collect all the domestic air tickets sold in the US and some additional management information and release the following three main tables: Coupon, Market, and Ticket. The Coupon table, which contains 880,384,622 rows in total, provides coupon-specific information for each domestic itinerary of the Origin and Destination Survey, such as the operating carrier, origin and destination airports, number of passengers, fare class, coupon type, trip break indicator, and distance. The Market table, which has 535,639,256 rows, contains directional market characteristics of each domestic itinerary of the Origin and Destination Survey, such as the reporting carrier, origin and destination airport, prorated market fare, number of market coupons, market miles flown, and carrier change indicators, and the Ticket table, which has 303,276,607 rows, contains summary characteristics of each domestic itinerary on the Origin and Destination Survey, including the reporting carrier, itinerary fare, number of passengers, originating airport, roundtrip indicator, and miles flown. Those thee tables share a set of common columns, i.e., primary-foreign key relationships in a database, and thus can be merged into one large table. Sometime airline names are changed so we use the unique identifiers assigned by the US governments rather than their names.}

\section{Final Feature Set in our Prediction}\label{sec:features}
The complete elements of $\mathbf{f}_{r,k}$ we use for our prediction are as follows so $\mathbf{f}_{r,k}$ is a 19-dimensional vector:
\begin{enumerate}
\item $f_{r,k,0}$: Average ticket price
\item $f_{r,k,1}$: Flight frequency
\item $f_{r,k,2}$: Delay ratio
\item $f_{r,k,3}$: Average delayed time in minutes
\item $f_{r,k,4}$: Flight cancel ratio
\item $f_{r,k,5}$: Flight divert ratio
\item $f_{r,k,6}$: Total number of fatal cases
\item $f_{r,k,7}$: Total number of serious accident cases
\item $f_{r,k,8}$: Total number of minor accident cases
\item $f_{r,k,9}$: Average aircraft size in terms of number of seats per flight
\item $f_{r,k,10}$: Average seat availability percentage which is not occupied by connecting passengers
\item $f_{r,k,11}$: In-degree of the source airport
\item $f_{r,k,12}$: In-degree of the destination airport
\item $f_{r,k,13}$: Out-degree of the source airport
\item $f_{r,k,14}$: Out-degree of the destination airport
\item $f_{r,k,15}$: PageRank of the source airport
\item $f_{r,k,16}$: PageRank of the destination airport
\item $f_{r,k,17}$: Ego network density of the source airport
\item $f_{r,k,18}$: Ego network density of the destination airport
\end{enumerate}

In our work, we optimize $f_{r,k,1}$ in each route to maximize the sum of market shares.


\section{Experimental Environments}
We introduce detailed environments we conducted our experiments on. We first describe software and hardware environments and then list detailed hyper-parameters.

Our detailed software environments are as follows:
\begin{enumerate}
    \item Ubuntu 18.04.1 LTS
    \item Python ver. 3.6.6
    \item Numpy ver. 1.14.5
    \item Scipy ver. 1.1.0
    \item Pandas ver. 0.23.4
    \item Matplotlib ver.3.0.0
    \item Tensorflow-gpu ver. 1.11.0
    \item CUDA ver. 10.0
    \item NVIDIA Driver ver. 417.22
\end{enumerate}

Our detailed hardware environments are as follows:
\begin{enumerate}
    \item Three machines with i9 CPU, each of which is equipped with 2-3 GPUs (GTX 1080 Ti).
\end{enumerate}

To train the three market share prediction models, Model1/2/3, we use the mini-batch size of 2,048 and a learning rate of 1e-4 which decays with a ratio of 0.96 every 100 epochs. We train 1,000 epochs for each model and use the Xavier initializer for initializing weights and the Adam optimizer for updating weights.

For market influence maximization, We have several hyper-parameters, $\bar{\beta}_{0}$, $\gamma_0$, $\lambda$, $\alpha$ and so on . All hyper-parameter configurations are already mentioned in the main paper.

\end{document}